\documentclass[10pt,twocolumn,letterpaper]{article}

\usepackage[final,applications]{wacv}      

\usepackage{graphicx}
\usepackage{amsmath}
\usepackage{amssymb}
\usepackage{booktabs}
\usepackage{cite}
\usepackage{algorithm}
\usepackage[noend]{algpseudocode}
\usepackage{xcolor, soul}
\usepackage{multirow}
\usepackage{caption}
\usepackage{subcaption}
\usepackage{wrapfig}
\usepackage{multirow}

\usepackage{pdflscape}
\usepackage{supertabular}
\usepackage{multicol}

\algrenewcommand\algorithmicforall{\textbf{foreach}}
\algrenewcommand\algorithmicindent{.8em}

\usepackage[pagebackref,breaklinks,colorlinks]{hyperref}

\usepackage[capitalize]{cleveref}
\crefname{section}{Sec.}{Secs.}
\Crefname{section}{Section}{Sections}
\Crefname{table}{Table}{Tables}
\crefname{table}{Tab.}{Tabs.}

\def\wacvPaperID{1572} 
\def\confName{WACV}
\def\confYear{2025}

\begin{document}

\title{Improving Medical Visual Representations via Radiology Report Generation}


\author{Keegan Quigley\\
MIT Lincoln Laboratory\\
Lexington, MA\\
{\tt\small keegan.quigley@ll.mit.edu}
\and
Miriam Cha\\
MIT Lincoln Laboratory\\
Lexington, MA\\
{\tt\small miriam.cha@ll.mit.edu}
\and 
Josh Barua\\
MIT Lincoln Laboratory\\
Lexington, MA\\
{\tt\small josh.barua@ll.mit.edu}
\and
Geeticka Chauhan\\
MIT CSAIL\\
Cambridge, MA\\
{\tt\small geeticka@mit.edu}
\and 
Seth Berkowitz\\
BIDMC\\
Boston, MA\\
{\tt\small sberkowi@bidmc.harvard.edu}
\and 
Steven Horng\\
BIDMC\\
Boston, MA\\
{\tt\small shorng@bidmc.harvard.edu}
\and
Polina Golland\\
MIT CSAIL\\
Cambridge, MA\\
{\tt\small polina@csail.mit.edu}
}
\maketitle

\begin{abstract}
   Vision-language pretraining has been shown to produce high-quality visual encoders which transfer efficiently to downstream computer vision tasks. 
   Contrastive learning approaches have increasingly been adopted for medical vision language pretraining (MVLP), yet recent developments in generative AI offer new modeling alternatives. This paper introduces RadTex, a CNN-encoder transformer-decoder architecture optimized for radiology. We explore bidirectional captioning as an alternative MVLP strategy and demonstrate that RadTex's captioning pretraining is competitive with established contrastive methods, achieving a CheXpert macro-AUC of 89.4\%. Additionally, RadTex's lightweight text decoder not only generates clinically relevant radiology reports (macro-F1 score of 0.349), but also provides targeted, interactive responses, highlighting the utility of bidirectional captioning in advancing medical image analysis.

\end{abstract}

\section{Introduction}
\label{sec:intro}

Automated medical image analysis has the potential to revolutionize healthcare diagnostics, but its efficacy is often constrained by the limited availability of annotated data. Acquiring high-quality annotations is time-consuming and requires expert knowledge, making fully supervised modeling approaches impractical. 

Recent advancements, including the release of large-scale weakly annotated datasets \cite{irvin_chexpert_2019,johnson_mimic-cxr-jpg_2019} and self-supervised medical vision-language pretraining (MVLP), have begun to address these limitations. Self-supervised techniques, such as contrastive learning exemplified by ConVIRT \cite{zhang_contrastive_2022} and CheXzero \cite{tiu_expert-level_2022}, have demonstrated substantial improvements in  image representation and impressive zero-shot capabilities using dual-encoder architectures.

When it comes to capturing textual semantics, however, contrastive learning is not the only option. Large language models (LLMs) have received widespread attention for their zero-shot capabilities on question-answering, comprehension \cite{brown2020language}, and reasoning tasks \cite{kojima2022large}. Generative language models use next token prediction tasks to capture complex meaning from unlabeled training sequences, and large-scale experiments have led to impressive results \cite{radford2018improving,raffel2020exploring}. 

Generative textual modeling has also been proposed as a method for visual-linguistic pretraining \cite{desai_virtex_2021}, with others theorizing more ``fine-grained" semantic encoding via captioning objectives \cite{yu_coca_2022}. Radiology is an application dominated by fine-grained semantics; slight differences in appearance can have substantially different diagnoses. While numerous approaches have adapted contrastive learning to address this challenge (see \cref{related:mvlp}), we propose that generative modeling objectives may induce learned representations that naturally capture these fine details in radiographs, improving MVLP.
Researchers recently explored similar hypotheses with MedPaLM M \cite{tu2024towards}, yet a lack of comparisons to contrastive pretraining, extensive general domain pretraining (780B tokens), and massive model sizes (12-562B parameters) obscure the comparative advantages of a generative approach to learning visual representations.

\begin{figure*}
    \centering
    \includegraphics[width=0.95\linewidth]{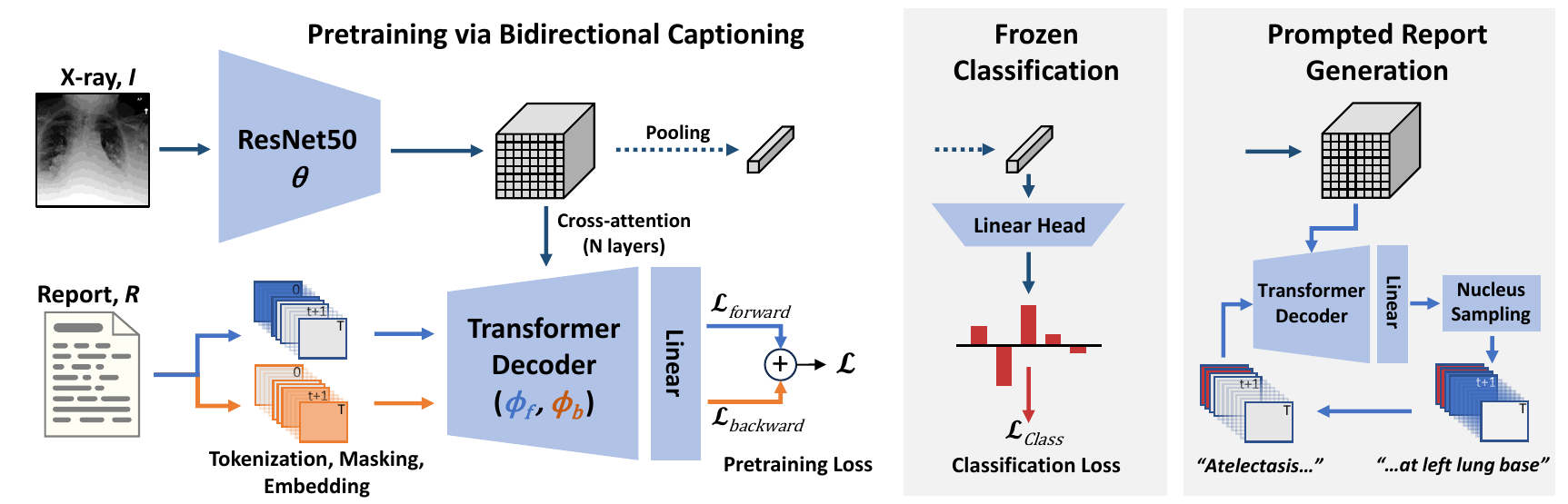}
    \caption{Overview of RadTex architecture and outputs, including pretraining, classification, and report generation in the \textit{Prompted} setting. We show that bidirectional captioning is an effective method of medical vision-language pretraining (left), exceeding contrastive learning performance on downstream visual tasks while also enhancing interpretability. After generative pretraining, the RadTex visual encoder is frozen and a linear head is trained to classify pathologies (center; \cref{sec:visual_downstream}). Furthermore, the entire pretrained model can be used directly for radiology report generation by sampling tokens from the output (right; \cref{sec:rrg}).}
    \label{fig:radtex-arch}
\end{figure*}

In this work, we introduce RadTex, a simple pretraining approach that leverages bidirectional captioning to meet the specific challenges of radiology. Our model extends the work of VirTex \cite{desai_virtex_2021} to radiology via domain-specific architectural changes and an extensive ablation study. With both smaller pretraining dataset and visual encoder sizes, RadTex outperforms CheXzero \cite{tiu_expert-level_2022} on three benchmarks for representation quality. 
This empirical result supports our hypothesis of fine-grained semantic encoding, opens a new avenue for MVLP research, and supports findings of large-scale multimodal language modeling studies.


Understanding the comparative strengths of generative pretraining relative to contrastive pretraining is the main objective of our work, but we also found surprising capabilities of the model for radiology report generation (RRG). While this field has received significant attention, we show that no fine-tuning is needed for RadTex to produce high-quality readable summaries of its findings, on par with leading \textit{RRG-dedicated} models. We further explore this capability by introducing prompting methods that elicit specific responses, enhancing clinical efficacy and demonstrating a pathway to improved radiologist-AI interaction. A model available for inference on a single GPU and capable of providing both diagnostic and interpretable outputs could unlock new safety-focused workflows in radiology.



\section{Related Work}
\label{sec:related}

Our work focuses on vision-language pretraining in the radiology domain, leveraging advancements in large language models and radiology report generation (RRG).

\subsection{Medical Vision-Language Pretraining}
\label{related:mvlp}

Prior to the advent of self-supervised deep learning approaches, copious annotations were needed to successfully train visual backbones such as ResNet and ViT \cite{he_deep_2015,dosovitskiy2021an}. Chauhan \etal \cite{chauhan_joint_2020} demonstrated the effectiveness of jointly encoding radiographs and radiology reports and encouraging similarity in their representations to address labeled data scarcity. ConVIRT \cite{zhang_contrastive_2022} similarly proposed contrastive loss \cite{chen_simple_2020} for vision-language tasks, which inspired the development of CLIP \cite{radford_learning_2021}, a state-of-the-art model for natural image-text representation learning. The pretrained CLIP model was fine-tuned for radiology in CheXzero  \cite{tiu_expert-level_2022}.


One challenge in contrastive pretraining has been the need to learn low-level semantics with global representations. GLoRIA
\cite{huang_gloria_2021} adopted local and global contrastive losses to learn low-level features, and MGCA \cite{wang2022multi} used prototypes to encourage disease-level clustering in representations. 
Another line of research explored augmenting contrastive MVLP with additional data sources such as PubMed \cite{mishra_improving_2023}, knowledge graphs \cite{zhang_knowledge-enhanced_2023}, and prior patient studies \cite{bannur2023learning}. 

Encoder-decoder architectures using transformers \cite{vaswani_attention_2017} have recently gained attention in MVLP. In the natural image domain, architectures like VirTex, Unicoder-VL, and ViLBERT have been successful for both pretraining and captioning \cite{lu_vilbert_2019,li_unicoder-vl_2020,zhou_unified_2020,desai_virtex_2021}. CoCa \cite{yu_coca_2022} achieved state of the art results by incorporating a textual decoder into a contrastive architecture. In the radiology domain, Bannur \etal \cite{bannur2023learning} utilized a masked language modeling (MLM) objective to augment their contrastive pretraining with a transformer decoder, and Yan \etal \cite{yan2022clinical} used MLM alongside other objectives to train an encoder-decoder but didn't make direct comparisons with contrastive approaches for MVLP. PRIOR \cite{cheng2023prior} took inspiration from CoCa, but used a prototype bank of sentences for modeling reconstruction.

Causal masking has been shown to be effective for pretraining language models \cite{brown2020language}. MedPaLM M \cite{tu2024towards} developed a question-answering system finetuned with next-token prediction objectives on a variety of tasks in language, imaging, and genomics, albeit at the billion-parameter scale. In comparison, we aim to develop a radiology-specific model with dataset and model scales similar to existing contrastive works, enabling a more concrete comparison of methods. 
Our architecture is most similar to VirTex \cite{desai_virtex_2021}, utilizing a CNN encoder and transformer decoder, with next-token prediction objectives. 

\subsection{Radiology Report Generation}

The image captioning task has been studied extensively in the natural image domain \cite{vinyals_show_2015,stefanini_show_2023}. However, captioning in the radiology domain poses unique challenges. To address challenges of linguistic and semantic consistency of generated reports, which are typically longer than natural image captions, Li \etal~\cite{li_hybrid_2018} and Pino \etal~\cite{pino_clinically_2021} have proposed using hybrid retrieval-generation models or templates. Endo \etal \cite{endo_retrieval-based_2021} proposed a report generation system that uses MVLP to improve retrieval.


Neural text generation, which generates text freely from an image, offers an alternative to contrastive zero-shot classification that does not require retrieval from a bank of templates or existing reports. Jing \etal \cite{jing_automatic_2018} and TIENet \cite{wang_tienet_2018} used LSTM-based generation, while $\mathcal{M}^2$ Trans \cite{miura-etal-2021-improving} and R2Gen \cite{chen_generating_2022} utilized transformers to generate relevant clinical information. Large language models have also been used to enhance report generation. CvT-212DistilGPT2 \cite{nicolson2023improving} finetuned a distilled LLM on MIMIC-CXR, and the generalist model MedPaLM M \cite{tu2024towards} demonstrated report generation with Pathways Language Model (PaLM) \cite{chowdhery2022palmscalinglanguagemodeling}.

\section{Methods}
\label{sec:methods}

We propose RadTex, a captioning-based pretraining approach for interpretable medical image analysis. The architecture is shown in \cref{fig:radtex-arch}. To implement the  model, we utilize the bicaptioning framework by Desai \etal~\cite{desai_virtex_2021}.

\subsection{Pretraining}
\label{sec:methods:pretraining}

Pretraining jointly optimizes an image encoder and textual decoder through bidirectional image captioning. The CXR image and paired radiology report inputs $[I, R]$, are first transformed into sequences for processing. A ResNet50 \cite{he_deep_2015} (denoted $\theta$) extracts visual features from $I$ and a linear projection creates a sequence of spatial image features $x_{vis}$. A tokenizer extracts tokens $w = \{w_0, ... , w_{T+1}\}$ from the report, and the tokens are embedded as the textual sequence $x_{text}$. 

A transformer decoder processes these two sequences, attending to $x_{text}$ through masked multiheaded self-attention and both $x_{text}$ and $x_{vis}$ via cross-attention in each of $L$ layers. 
The transformer architecture is duplicated for the backward captioning transformer, and weights are not shared. Attention masks when predicting the $t$-th token $w_t$ are applied to $w_{i\geq t}$ and $w_{i\leq t}$ for the forward and backward transformers, respectively.

Outputs from forward and backward transformers are passed through a single linear layer, which shares weights with the token embedding layers (following \cite{desai_virtex_2021,press2016using}), to predict token log-probabilities. We compute cross-entropy loss on these logits, minimizing negative log-likelihood of selecting the correct tokens as follows: 
\vadjust{\penalty\predisplaypenalty\vskip-\jot\relax}
\small
\begin{multline}
\mathcal{L} =  - \sum^{T+1}_{t=1} \log p(w_t | I, w_{i<t};\theta,\phi_f) \\
- \sum^{T}_{t=0} \log p(w_t | I, w_{i>t};\theta, \phi_b)
\label{equation:loss}
\end{multline}
\normalsize
Here, $\phi_f$ and $\phi_b$ represent the parameters for the forward and backward decoders. In this way, prediction of $t$-th token $w_t$ is conditioned on \{$w_{i<t}$, $x_{vis}$\} for forward captioning and \{$w_{i>t}$, $x_{vis}$\} for backward captioning. 

With the encoder-decoder pretraining formulation and captioning pretraining, we adopt a cross-attention calculation between token embeddings and image regions which is inherently \textit{local}. 
Crucially, this differs from the standard contrastive objective where similarity is encouraged between the \textit{global} representations of image and text. 
Like CoCa \cite{yu_coca_2022}, we hypothesize that our setup might lead to advantages for fine-grained visual tasks.


\subsection{Unprompted and Prompted Captioning}
\label{sec:methods_captioning}

The pretrained encoder-decoder architecture can also generate reports by predicting the probability of a token $w_t$ based on visual data and either preceding tokens \{$w_{i<t}, x_{vis}; \phi_f$\} or successive tokens \{$w_{i>t}, x_{vis}; \phi_b$\}. We employ autoregressive captioning to iteratively build upon an initial sequence by adding one token at a time.
Tokens are sampled with nucleus sampling \cite{holtzman_curious_2020}, which we found to work better than VirTex's beam search. 

An initial token sequence $P$ for autoregressive decoding can optionally be non-empty. We evaluate with an empty starting sequence ($P = \emptyset$), termed \textit{Unprompted}, and non-empty $P$ to examine the effectiveness of ``prompting" the decoder with textual indicators. When prompting, we limit generation to a single sentence to prevent off-topic generation, and append sentences in the case of multiple prompts. The generalized approach to autoregressive prompted captioning is detailed in \cref{app:prompted-captioning}.

We propose two prompting strategies. 
The first, \textit{Prompted} captioning, begins with adding a word or phrase with $N$ tokens to $P$, prompting the model to run forward captioning to generate outputs $w_{i>N}$. 
The second, \textit{Iterative Prompted} captioning, starts with the \textit{Prompted} procedure then re-initializes $P$ with the output sequence and performs backward captioning. As such, the model generates tokens before \textit{and} after the initial prompt. Given the typically rigid linguistics of radiology reports, allowing for forward and backward captioning on the same prompt may generate text that more completely details a particular finding.

\section{Experimental Settings}
\label{sec:experiments}

\subsection{Datasets}
We use MS-COCO \cite{lin_microsoft_2014} and MIMIC-CXR \cite{johnson_mimic-cxr_2019,johnson_mimic-cxr-jpg_2019} for pretraining, and transfer our visual encoder to a variety of downstream datasets. Dataset details are described here:\\

\noindent \textbf{MS-COCO }\cite{lin_microsoft_2014} comprises natural images each paired with captions. For pretraining the RadTex model, we use the official 2017 split with 118K images. \\

\noindent \textbf{MIMIC-CXR} \cite{johnson_mimic-cxr_2019,johnson_mimic-cxr-jpg_2019} provides 377,110 chest X-rays (CXRs) with paired reports and CheXpert labels. For pre-training, we use the \textit{Findings} and \textit{Impression} sections and both frontal and lateral MIMIC-CXR-JPG images. Reports are preprocessed to remove references to prior studies (see \cref{sec:hallucinations}) in the training and validation sets (train/val/test: 368,960/2991/5159). For  downstream classification, we use Pathology9 \cite{liao_multimodal_2021}, containing 9 of the 14 original CheXpert labels with over 100 test examples. \\

\noindent \textbf{CheXpert} \cite{irvin_chexpert_2019} includes 224,316 chest radiographs from Stanford Hospital. Following the official split (223,414 training examples), our experiments focus on classifying five competition pathologies: Atelectasis, Cardiomegaly, Consolidation, Edema, and Pleural Effusion. 
\\

\noindent \textbf{Edema Severity} \cite{horng_deep_2021} is derived from MIMIC-CXR, grading 7,390 radiographs for pulmonary edema severity on a 0-3 scale (0: none, 1: vascular congestion, 2: interstitial edema, 3: alveolar edema). While most labels derive from regular expressions (regex), the test set ($n=141$) uses consensus labels from radiologists. This test set is unseen during MIMIC-CXR-JPG pretraining. \\

\noindent \textbf{RSNA Pneumonia} \cite{shih_augmenting_2019} comprises approximately 30,000 radiographs from NIH CXR-8 \cite{wang_chestx-ray8_2017}, labeled for pneumonia. \\

\noindent \textbf{COVIDx} \cite{pavlova_covidx_2022} is an updated version of the multinational COVIDx-CXR3 data. The dataset contains COVID and non-COVID labels for 30,386 images. We report on their official, class balanced, 400-image test set, and use 5\% of the training set for validation set, split by patient identifier.



\subsection{Implementation Details}


We monitor validation loss during pretraining and use the best model checkpoint for downstream tasks. For all downstream tasks in \cref{section:ablations,sec:visual_downstream}, we extract the visual encoder from the pretrained model and freeze model weights, training only a classification head for the pooled visual encoder features. All reported downstream results are the mean and standard deviation of three trials, unless otherwise stated. Additional implementation details can be found in \cref{sec:extra_implementation_details}. We will make our code available upon publication to ensure reproducibility.

\section{Results and Discussion}

We perform an extensive set of experiments that optimize pretraining, evaluate bidirectional captioning for MVLP, and demonstrate interpretable report generation in multiple RadTex configurations.

\subsection{Optimizing RadTex for Radiology}
\label{section:ablations}
Prior to running ablations, we adjust key settings to better suit the general-domain bicaptioning model (i.e. VirTex) to the radiology domain.
We increase the context length of the transformer decoder from 30 to 170 tokens, covering 99\% of the MIMIC-CXR \textit{Findings} and \textit{Impression} sections. Following Chauhan \etal \cite{chauhan_joint_2020}, we use the SciBERT tokenizer vocabulary \cite{beltagy_scibert_2019} instead of SentencePiece \cite{kudo_sentencepiece_2018}. 
And lastly, we apply domain-specific augmentations to match the expected image orientations (see \cref{sec:extra_implementation_details}).


\subsubsection{Pretraining Ablations}
With these settings as the baseline we perform ablations on pretraining, evaluating ablation performance via linear fine-tuning on the Pathology9 classification task. 

\begin{table}[t]
    \begin{tabular}{@{\extracolsep{\fill}} clrr }
    \toprule[1pt]
            Ablation & Parameter & Path9 AUC & $\Delta$AUC\\
            \midrule[1pt]
            \multirow{3}*{
                \begin{tabular}{c}
                    Pretraining\\
                    Steps
                \end{tabular}} & 50K & $79.3_{\pm 0.1}$ & ---\\
            & \textbf{100K}  & $\mathbf{80.1_{\pm 0.1}}$ & $\mathbf{+0.8}$ \\
            &200K  & $79.8_{\pm 0.1}$ & $+0.5$ \\
            \midrule[0.25pt]
            \multirow{3}*{
                \begin{tabular}{c}
                    Transformer\\
                    Layers $L$
                \end{tabular}}& 1 Layer & $79.3_{\pm 0.1}$ & ---\\
            &\textbf{2 Layers}  & $\mathbf{80.1_{\pm 0.1}}$ & $\mathbf{+0.8}$\\
            &4 Layers & $79.7_{\pm 0.1}$ & $+0.4$ \\
            \midrule[0.25pt]
            \multirow{3}*{
                \begin{tabular}{c}
                    Tokenizer\\
                    Vocab
                \end{tabular}}& \textbf{SciBERT} & $\mathbf{79.3_{\pm 0.1}}$ & ---\\
            &COCO SP  & $79.4_{\pm 0.0}$ & $+0.1$ \\
            &MIMIC SP & $79.0_{\pm 0.1}$ & $-0.3$ \\
            \midrule[0.25pt]
            \multirow{2}*{
            \begin{tabular}{c}
                 Captioning\\
                 Loss
            \end{tabular}
            }
            & \textbf{Bidirectional}  & $\mathbf{79.3_{\pm 0.1}}$ & ---\\
            & Forward  & $77.4_{\pm 0.2}$ & $-1.9$\\
            \midrule[0.25pt]
            \multirow{2}*{Prior Refs.}
            & \textbf{Removed}  & $\mathbf{79.3_{\pm 0.1}}$ & ---\\
            &  Present  & $78.7_{\pm 0.1}$ & $-0.6$  \\
            \bottomrule
    \end{tabular}
    \caption{Results of RadTex pretraining ablation study, measured by linear fine-tuning performance (AUC) on Pathology9. Average of three trials is reported with $\text{Mean}_{\pm\text{SD}}$. 
    Significant improvements over the baseline for each ablation are adopted in the final model, and those selected parameters are \textbf{bolded}.
    }
    \label{tab:ablations}
\end{table}

We increase the number of \textbf{pretraining steps} of our baseline model from 50K to 100K and 200K, In \cref{tab:ablations}, we find that both improve downstream classification performance, but the model with 100K steps performs the best. 

Following the ablation study in VirTex \cite{desai_virtex_2021}, we experiment with additional \textbf{transformer layers} in the decoder. Notably, we observe significant improvement with 2 layers.

One of the key settings changed for the baseline model was the \textbf{tokenizer vocabulary}. We test both the COCO SentencePiece (used in VirTex) and MIMIC SentencePiece computed from the MIMIC training corpus. Despite an in-domain vocabulary, the MIMIC SentencePiece model performs worse when it comes to downstream classification. The COCO model performs slightly better than our baseline (\cref{tab:ablations}) despite the out-of-domain vocabulary, but the margin is insignificant ($p>0.05$ in a one-sided t-test).

Sentences in radiology reports often follow similar syntactic and semantic patterns when describing CXRs, so we investigate whether unidirectional \textbf{captioning loss} is a sufficient pretraining objective. By removing the second term in \cref{equation:loss}, we reduce the bidirectional model to \textit{Forward Only}. We find that the bicaptioning approach is still necessary to fully leverage the textual information.

As part of pretraining dataset preparation, we remove sentences in the MIMIC-CXR training corpus containing the phrases ``prior" and ``compar." We hypothesize that the presence of \textbf{prior references} trains the model to hallucinate prior studies, of which it could have no knowledge. Removing the phrases is shown to both improve downstream classification (\cref{tab:ablations}) and reduce the number of references to prior reports (\cref{sec:hallucinations}).

\subsubsection{Final RadTex Model}
Our final RadTex model incorporates baseline settings and findings from the ablation study (100K steps, 2 layers, SciBERT vocabulary, bidirectional captioning, priors removed). This model achieves a Pathology9 AUC of $80.5$\%, which exceeds all other ablations. We additionally pretrain RadTex on MS-COCO before MIMIC-CXR for better weight initialization. We report these results as RadTex/C+M, or simply RadTex, while the model trained solely on MIMIC-CXR is denoted as RadTex/M.



         

\subsection{Visual Downstream Tasks}
\label{sec:visual_downstream}

\begin{figure*}[t]
    \centering
    \begin{minipage}{0.68\textwidth}
        \centering
        \includegraphics[width=0.99\textwidth]{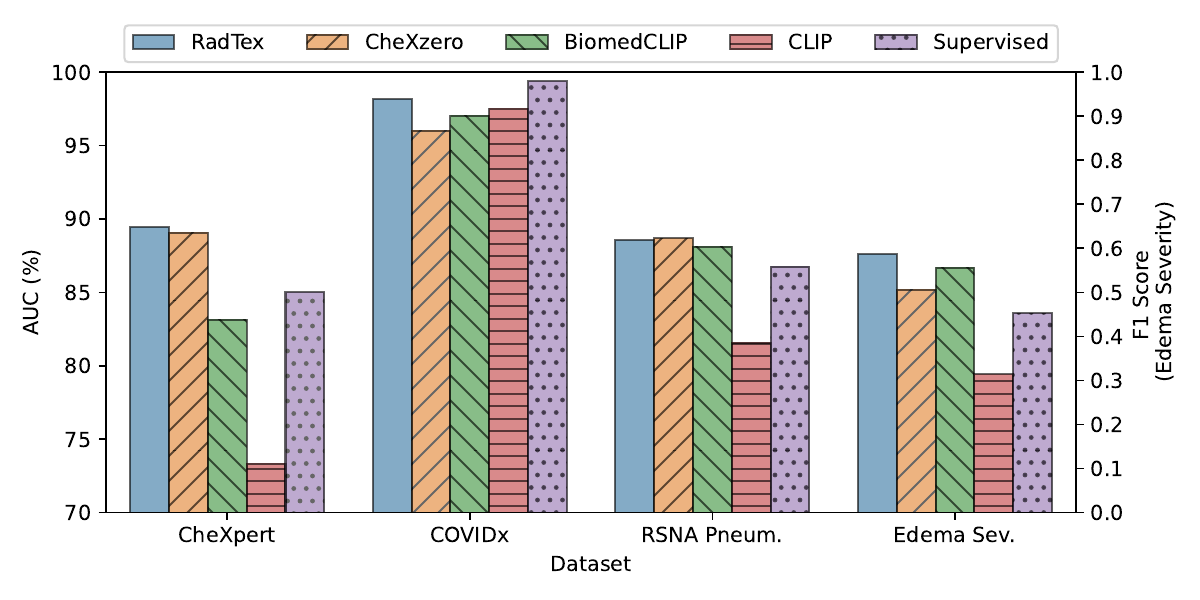} 
        \caption{Bar plot showing linear classification results. RadTex is competitive with CheXzero and other methods across multiple downstream classification tasks. RadTex results are for RadTex/C+M pretraining. Each model's visual backbone is frozen and a linear layer is trained in three separate trials. We display mean results over three random trials. See \cref{app:vision_encoder_results} for more details, including standard errors.}
    \label{fig:classification}
    \end{minipage}\hfill
    \begin{minipage}{0.3\textwidth}
        \centering
        \includegraphics[width=0.95\linewidth]{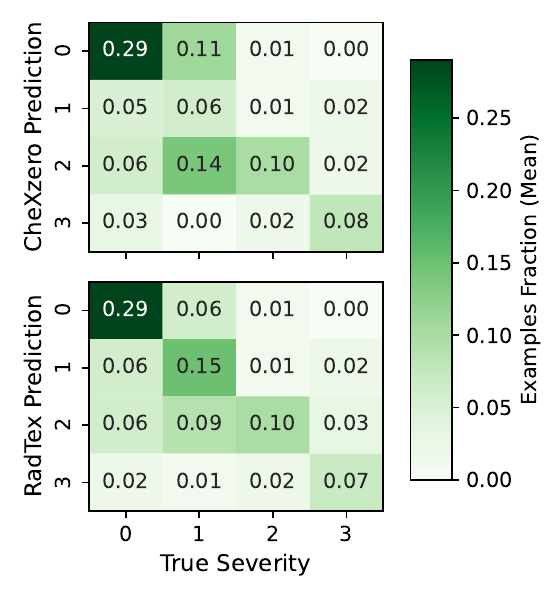}        
        \caption{Confusion matrices showing multi-level grading performance for RadTex (bottom) and CheXzero (top), on the EdemaSeverity task. Values represent the mean proportion over three random trials.}%
    \label{fig:edema_severity}
    \end{minipage}
\end{figure*}


We evaluate the quality of our MVLP technique on four downstream classification tasks: CheXpert (multilabel  on five competition pathologies), COVIDx (binary), RSNA pneumonia detection (binary), and Pulmonary Edema Severity classification (ordinal). These tasks probe both high- and low-level semantic feature encodings.



\subsubsection{Transferable Representations from Bicaptioning}

For the most direct comparison of visual encoder quality, we evaluate models in a \textit{frozen} linear classification setup, where a randomly initialized linear classification head is added to a pretrained encoder, and only the linear head is trained on available data. 

We obtain the final visual encoder weights from CheXzero (ViT-B/32), BiomedCLIP (ViT-B/16), and CLIP (ResNet50) to test in these settings. CheXzero, initialized from CLIP and pretrained on MIMIC-CXR reports, is considered the closest contrastive comparison in terms of visual encoder size and pretraining datasets. BiomedCLIP is trained on PMC-15M with 15 million image-text pairs from PubMed (an alternative data source for this domain), and CLIP, pretrained on 400M image-text pairs serves as a natural domain baseline. For an additional baseline, we train a randomly initialized ResNet50 end-to-end in supervised fashion on the available labels for each task.

We find that RadTex performs the best against the frozen encoder benchmarks in three out of four experiments (\cref{fig:classification,tab:visual_experiments}), showcasing the power of generative pretraining. On the COVIDx task, which involves binary classification on a condition not covered in the pretraining data, RadTex achieves the highest score among the pretrained models, demonstrating RadTex's capability to extract fine-grained evidence from CXRs and its generalizability to previously unseen data. Supervised modeling exceeds all frozen visual encoders on COVIDx, but in an additional experiment where all encoders are fine-tuned end-to-end, RadTex achieves the highest score across all baselines.


Like RadTex, MedPaLM M trained a visual encoder on generative objectives, but did so across multiple textual datasets and with billion-parameter scale models. They reported few classification results and did not make their code available, so direct comparison to RadTex is difficult. 
However, they report a macro-AUC of 79.09\% on the CheXpert competition pathologies (Table 2 of Tu \etal \cite{tu2024towards}). We replicated their classification experiment with a frozen RadTex encoder, finding a macro-AUC of 84.07\%, suggesting that much smaller, domain-specific models may outperform larger, generalist ones in specialized tasks.
\subsubsection{Data Efficient Transfer Learning}
Assessing data efficiency of models is crucial in the medical domain due to the difficulty in gathering large, annotated datasets. In \cref{tab:chexpert_partial}, we evaluate RadTex under conditions of limited labeled data and compare against contrastive models like GLoRIA \cite{huang_gloria_2021} and MGCA \cite{wang2022multi}, utilizing their ResNet50 versions for direct comparison. 
Despite GLoRIA and MGCA's complex semantic encoding methods, they fall short of CheXzero. This could be due to the larger ViT-B  visual encoder of CheXzero (88M params), or its additional pretraining on natural domain data. 
Surprisingly, the RadTex ResNet50 (24M params) exceeds CheXzero's ViT-B at all levels of data availability, speaking to the effectiveness of bidirectional captioning pretraining. A comparison with the original VirTex architecture highlights the value of our domain-specific modifications, detailed in \cref{section:ablations}.

\begin{table}[t]
\small
\begin{tabular}{p{0.0cm}llp{1.05cm}p{1.05cm}p{1.05cm}}
\toprule
                 & & Visual  & \multicolumn{3}{c}{CheXpert AUC (\%)}\\
                 & & Enc. & 1\%    & 10\%     & 100\%  \\
                 \midrule
& Random Init* & RN50 & $71.9_{\pm2.6}$ & $82.3_{\pm3.5}$ & $85.0_{\pm4.7}$ \\
\midrule
{\multirow{6}{*}{\rotatebox[origin=c]{90}{Contrastive}}} 
& OpenAI CLIP & RN50  & $59.8_{\pm0.4}$ & $72.4_{\pm0.6}$ & $73.3_{\pm0.2}$  \\
& ConVIRT & RN50  & $85.9$ & $86.8$ & $87.3$ \\
& GLoRIA & RN50  & $86.6$ & $87.8$ & $88.1$ \\
& MGCA & RN50  & $87.6$ & $88.0$ & $88.2$ \\
& BiomedCLIP & ViT-B  &$77.2_{\pm0.9}$ & $82.1_{\pm2.6}$ & $83.1_{\pm0.5}$ \\
& CheXzero  & ViT-B   &  $88.9_{\pm0.6}$ & $89.1_{\pm0.2}$  & $89.0_{\pm0.4}$ \\
\midrule
{\multirow{3}{*}{\rotatebox[origin=c]{90}{Cap.}}} 
& VirTex/C+M & RN50 & $86.6_{\pm0.8}$ & $86.7_{\pm0.7}$ & $87.3_{\pm0.2}$ \\
& RadTex/M  & RN50        & $88.4_{\pm0.2}$ & $89.2_{\pm0.5}$ & $89.0_{\pm0.4}$  \\
& RadTex/C+M   & RN50   &$\mathbf{89.2_{\pm0.4}}$ & $\mathbf{89.6_{\pm0.1}}$ & $\mathbf{89.4_{\pm0.1}}$ \\ \bottomrule \\
\end{tabular}
\caption{CheXpert competition linear classification results after linear fine-tuning on limited labeled data (1\%, 10\%, and 100\% of the CheXpert training set). RadTex's visual encoder exceeds performance of CheXzero's ViT-B, despite a much smaller model size. $\text{Mean}_{\pm\text{SD}}$ AUC across 3 trials presented. ConVIRT, GLoRIA, and MGCA results from \cite{zhang_contrastive_2022}, \cite{huang_gloria_2021}, and \cite{wang2022multi}, respectively. 
}
\label{tab:chexpert_partial}
\end{table}

\subsubsection{Fine-grained Pathology Classification}
Precise diagnosis for determining a course of treatment requires distinguishing between closely related conditions. We theorize that successful analysis of subtle differences in pathology requires fine-grained visual encoding (\cref{sec:methods:pretraining}). The EdemaSeverity task, multi-level grading of a single pathology, provides insights into the ability of the model to make such distinctions.
We analyze EdemaSeverity confusion matrices for RadTex and CheXzero in \cref{fig:edema_severity}. 
The models perform similarly when classifying the most severe cases of alveolar edema (level 3), but we find that RadTex outperforms CheXzero in differentiating between mild and moderate edema (levels 1 and 2). This distinction is more challenging, according to Horng \etal \cite{horng_deep_2021}, suggesting advantages for fine-grained classification with RadTex.

\newcommand{\ctext}[3][RGB]{%
  \begingroup
  \definecolor{hlcolor}{#1}{#2}\sethlcolor{hlcolor}%
  \hl{#3}%
  \endgroup
}

\newcommand{\red}[1]{\ctext[RGB]{255,191,191}{#1}}
\newcommand{\green}[1]{\ctext[RGB]{210,231,214}{#1}}
\newcommand{\yellow}[1]{\textcolor{black}{\ctext[RGB]{255,241,146}{#1}}}

\begin{table*}[t]
\small
\centering
     \begin{tabular}{ c p{0.38\linewidth} p{0.33\linewidth}}
     \toprule
     CXR Image & Ground Truth & RadTex Generated Report \\
     \midrule
     \raisebox{-1.05\totalheight}{\includegraphics[width=0.2\textwidth]{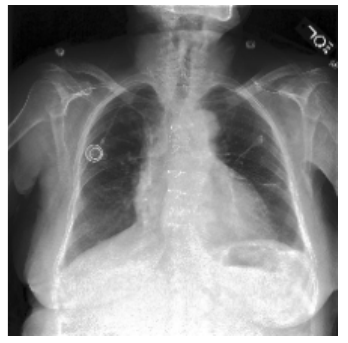}}
        &  Right port catheter tip ends in the 
mid svc. Sternal wires are intact and midline. There are small bilateral pleural effusions, slightly larger on the left than on the right. The cardiac silhouette is moderately enlarged. There is mild engorgement of the pulmonary vasculature. There has been improvement in the previously noted pulmonary edema with minimal residual edema. There is plate-like atelectasis seen in the left base. There is no consolidation or pneumothorax. Number. Small bilateral pleural effusions. Number. Improvement in pulmonary edema.
      & 
 \red{Left-sided} \green{central venous catheter terminates in the mid svc.} \red{Overall cardiac size is top normal.} Aortic knob is noted. \green{Patient is status post median sternotomy} \yellow{and intact median sternotomy} with \green{intact median sternotomy wires.} \yellow{Mediastinal contours are stable} in appearance. \yellow{Residual} \green{right pleural effusion with airspace opacity seen on the left is present}, likely \green{atelectasis.} \green{Atelectasis is seen at the left lung base. No pneumothorax. If there is a small left pleural effusion.} \\

      \bottomrule
      \end{tabular}
      \caption{RadTex/C+M captioning on a test-set CXR, comparing radiologist-written Ground Truth and \textit{Unprompted} Report. Agreement of generated report with ground truth is highlighted. Key: \green{GT Agreement} \red{GT Disagreement} \textcolor{gray}{\yellow{Irrelevant Info}}}
      \label{tab:cap_examples}
\end{table*}

\subsubsection{Captioning vs. Contrastive MVLP}
\begin{table*}[th]
\small
\centering
    \begin{tabular*}{\linewidth}{@{\extracolsep{\fill}} lrl|cc|ccc }
    \toprule
            &&& \multicolumn{2}{c|}{Textual Similarity} & \multicolumn{3}{c}{Clinical Efficacy (CE)} \\
             & && BLEU-2 & BERTScore\textdagger & CheXpert & CheXbert\textdagger & RadGraph\textdagger \\
            Model & Params (M) & Method & & & \footnotesize{macro-F1} & \footnotesize{Cosine Sim.} & \footnotesize{F1} \\ 
            \midrule
            Random Report Retrieval & --- & Retrieval & 0.089 & 0.213 & 0.177 & 0.166 & 0.048 \\
            $\mathcal{M}^2$ Trans \cite{miura-etal-2021-improving} & 28 & Generative & -- & 0.227 &  0.304 & 0.268 & 0.110\\
            R2Gen \cite{chen_generating_2022} & 86 & Generative & \textbf{0.218} & 0.186 & 0.276 & 0.204 & 0.057\\
            CXR-RePaiR\cite{endo_retrieval-based_2021} & 151 & Retrieval & 0.069 & 0.191 & 0.256 & \textbf{0.379} & 0.091\\
            \midrule
            RadTex Unprompted & 360 & Generative &  0.100 & 0.261 & 0.289 & 0.259 & 0.096 \\
            RadTex Prompted  & 360 & Generative &  0.069 & 0.262 &  \textbf{0.349} & 0.336 & 0.098 \\
            RadTex Iterative Prompt & 360 &Generative  &   0.082 & \textbf{0.271} & \textbf{0.349} & 0.333 & \textbf{{0.112}} \\
            \midrule
            MedPaLM M \cite{tu2024towards} & 84000 & Generative & -- & -- & \textit{0.398} & -- & \textit{0.267} \\
            \bottomrule

    \end{tabular*}
    \caption{Comparison of radiology report captioning techniques on a range of metrics. For BLEU-2 and CheXpert F1 scores, R2Gen and CXR-RePaiR compare to ground truths with both \textit{Findings} and \textit{Impression}, while $\mathcal{M}^2$ Trans uses just \textit{Findings}. We include MedPaLM M for reference, but consider it an unequal comparison, given that the model was provided with the \textit{Indication} section along with the image as priors. \textdagger Following \cite{yu_evaluating_2022} and using only \textit{Impression} section for ground truth. }
    \label{tab:captioning}
\end{table*}


We propose two hypotheses for the success of bidirectional captioning pretraining. First, while traditional contrastive learning primarily captures high-level global semantics of images and text, diagnostic tasks often require finer-grained understanding of images \cite{wang2022multi}. RadTex, using a token-level loss function and bidirectional causal masking, leverages the dense semantics of text to capture fine-grained image features important for effective diagnostics.
Second, contrastive learning may inadvertently classify similar examples as dissimilar, potentially leading to false negatives \cite{huynh2022boosting}, a critical issue given the semantic similarity often displayed in radiographic images. Captioning models, in contrast, can encode these hard negatives without artificially enforcing dissimilarity. 
These hypotheses are further supported by RadTex's superior performance on COVIDx and Edema Severity tasks, where out-of-distribution examples and severity differentiation challenge the models.




\subsection{Radiology Report Generation}
\label{sec:rrg}
An important consideration for visual analysis methods in radiology is their interpretability. Given the substantial variability in pathology appearance and overlap between normal and pathological findings, diagnosis based on imaging alone can sometimes be impossible. While the primary focus of our work is improving MVLP with captioning objectives, we must also consider how RadTex can assist in the interpretation of radiographs. 


Contrastive pretraining can deliver interpretation through zero-shot classification but it requires the careful formulation of binary textual inputs, a restrictive setup that is not conducive to open-ended diagnosis assistance. On the other hand, RadTex's inherent ability to perform radiology report generation provides an interpretable zero-shot output which fits with existing radiology workflows. We find surprising RRG performance, and extend our contributions by investigating captioning in \textit{Unprompted}, \textit{Prompted}, and \textit{Iterative Prompted} settings, as described in \cref{sec:methods_captioning}, with RadTex/C+M pretraining.


In Prompted Captioning, any word or sequence of tokens can be used as a prompt, creating opportunities for including clinical priors in captioning tasks, or potentially eliciting insights into rare or abnormal pathologies.
However, for optimal clinical efficacy and to demonstrate a fully automated technique, we use the outputs from our Pathology9 classifier with thresholds selected by Youden's indices \cite{youden1950index} on validation set for prompt selection. The model is prompted with only the positive pathologies, and their outputs are concatenated to form the final report.


\subsubsection{Caption Quality Evaluations}
\cref{tab:cap_examples} presents an example \textit {Unprompted} report generated by RadTex, demonstrating the textual outputs that might assist a radiologist. The report successfully identified a catheter position, detected bilateral pleural effusion, and noted atelectasis while confirming an absence of pneumothorax. However, it incorrectly determined the side of the catheter and slightly misrepresented the cardiac silhouette. 
We shade the generated words based on their agreement with the ground truth report. 

For quantitative comparisons to existing RRG methods, we measure textual similarity and clinical efficacy (CE). BLEU-2 and BERTScore \cite{zhang_bertscore_2019} are common natural language generation (NLG) metrics for syntactic and semantic similarity, but fail to describe the diagnostic accuracy of reports. Clinical efficacy is evaluated with CheXpert macro-F1 scores, CheXbert embedding cosine similarities between prediction and ground truth \cite{smit2020chexbert}, and RadGraph F1 \cite{yu_evaluating_2022}, which converts reports into knowledge graphs and computes overlap with ground truth.

We generate captions for 5,159 MIMIC-CXR test set radiographs. \cref{tab:captioning} compares RadTex to other RRG methods: $\mathcal{M}^2$ Trans \cite{cornia_meshed-memory_2020,miura-etal-2021-improving} and R2Gen \cite{chen_generating_2022} are two top-performing transformer-based methods, and CXR-RePaiR \cite{endo_retrieval-based_2021} is cited as a report retrieval technique.
MedPaLM M results are provided, although  Tu \etal \cite{tu2024towards} added the \textit{Indication} section to model context for RRG, giving the model additional priors which boost performance and preclude direct comparison.

In \cref{tab:captioning}, RadTex performs comparably to other transformer-based RRG methods: it falls between R2Gen and $\mathcal{M}^2$ Trans on CE when unprompted, but exceeds  $\mathcal{M}^2$ Trans when prompted. Retrieval-based CXR-RePaiR performs best on CheXbert similarity, which could be due to direct retrieval of common report phrases from a sentence bank. Textual similarity metrics are split, with RadTex falling behind R2Gen on BLEU-2, but exceeding other methods on BERTScore. BLEU-2, a measure of n-gram overlap, may be negatively affected by the use of nucleus sampling, which promotes diversity in sampled tokens.

Observations of RadTex reports by authors and an attending radiologist suggest that prompting with positively-classified classes may also improve fine-grained explainability of the reports. However, the radiologist also pointed out instances where RadTex generated reports that connected unrelated concepts, such as associating pathologies with no causal link. When we test prompting with negative-classified pathologies, RadTex shows difficulty negating the initial prompt in continued generation. With this setup, CheXbert embedding similarity scores fall to 0.040, indicating a sensitivity in clinical accuracy to prompt formulation. 
Given impressive fine-grained visual encoder results, we believe that there is potential for accurately transcribing these details in RRG, but further research into prompting and sampling strategies is needed. Scaling model and pretraining dataset sizes may also help improve general reasoning, as evidenced by recent trends in LLM research \cite{radford2018improving,tu2024towards}.

\subsubsection{Hallucinations}
\label{sec:hallucinations}

A widely acknowledged challenge of generative AI systems is a tendency to issue plausible but incorrect generations, often reflective of patterns in training data\cite{rohrbach2018object}. It is vital to measure hallucinations in medical generative AI given the potential for misdiagnosis, yet pathology hallucinations can be difficult to quantify given the frequent disagreement between experts on ground truth labels\cite{horng_deep_2021}. To start, we measure a more concrete hallucination: references to prior reports that are unbeknownst to the model.  

Because radiograph-report pairs are a temporal snapshot in the course of a patient's care, radiologists often reference previous studies in their reports. In experiments, we used a regex-labeler to identify prior references in our generated reports, and found that 43.0\% contained prior reference hallucinations. Using the same labeler, we removed sentences from ground-truth reports with prior references and re-started pretraining, yielding a model with a 4.0\% reduction in clinical efficacy (CheXpert F1) but a 99.8\% reduction in these hallucinations. A recent study by Ramesh \etal \cite{ramesh_improving_2022} proposed re-writing ground-truth reports containing priors using GPT-3, a preprocessing approach which could reduce impact on clinical efficacy in the future.

To measure pathology hallucination, we study CheXpert-derived labels of our generated reports for the MIMIC-CXR test set and compute per-pathology precision, recall, and F1 scores (see \cref{tab:per_pathology}). Precision, in particular, may indicate a model's tendency to hallucinate about non-existent pathologies, and we observe a strong positive correlation between RadTex precision score and a pathology's frequency in the training data (\cref{fig:per_pathology_captioning}). We  find similar trends for other transformer-based RRG methods and metrics. This suggests that improved class balance in pretraining data could reduce pathology hallucinations.

\subsection{Clinical Application Assessment}


The RadTex approach demonstrates success by addressing two major challenges in automated assistance for clinical radiology: fine-grained semantic encoding, and a need for interpretability. These challenges define the radiology application, and we argue that generative pretraining addresses both simultaneously. A clinical workflow could involve RadTex generating a diagnostic summary and preliminary report for radiologists to review and validate, accelerating the diagnostic process---on a single Nvidia V100 GPU, RadTex can generate a report in 0.19s. However, work remains before RadTex is ready for clinical use.

Recent studies have shown that CXR analysis algorithms contain biases that create negative social impact. Models can take ``shortcuts" in the assessment of patient health, biasing their predictions on demographics rather than pathological findings \cite{seyyed2021underdiagnosis,yang2024limits}. Closer evaluation of RadTex performance in minority subpopulations will be critical for the safe deployment of the model in clinical environments, and we plan to address these challenges in future work.

Beyond radiology, other application areas with similar challenges stand to benefit from our work. Geospatial analysis and autonomous scene understanding require localized attention to fine-grained elements in the visual field. Remote sensing, in particular, would be an excellent fit due to the common need for reports based on visual findings.


\section{Conclusion}
\label{sec:conc}
In this paper, we demonstrate the efficacy of bidirectional captioning as a pretraining strategy for interpretable medical image analysis. Our model, RadTex, not only yields competitive performance against contrastive methods in visual downstream tasks but also exhibits evidence of improved fine-grained visual encoding. A distinct advantage of RadTex is its capability to generate radiology reports, providing levels of interpretability not achieved by other contrastive learning-based pretraining methods. Additionally, we introduce a flexible prompting mechanism, showcasing its potential to improve clinical efficacy. 




\section*{Acknowledgements}
\noindent DISTRIBUTION STATEMENT A. Approved for public release. Distribution is unlimited.
This material is based upon work supported by the Department of the Air Force under Air Force Contract No. FA8702-15-D-0001. Any opinions, findings, conclusions or recommendations expressed in this material are those of the author(s) and do not necessarily reflect the views of the Department of the Air Force.
\\\\
\noindent © 2024 Massachusetts Institute of Technology.
\\\\
\noindent Delivered to the U.S. Government with Unlimited Rights, as defined in DFARS Part 252.227-7013 or 7014 (Feb 2014). Notwithstanding any copyright notice, U.S. Government rights in this work are defined by DFARS 252.227-7013 or DFARS 252.227-7014 as detailed above. Use of this work other than as specifically authorized by the U.S. Government may violate any copyrights that exist in this work.
\newpage
{\small
\bibliographystyle{ieee_fullname}
\bibliography{egbib}

\begin{thebibliography}{10}\itemsep=-1pt

\bibitem{bannur2023learning}
Shruthi Bannur, Stephanie Hyland, Qianchu Liu, Fernando Perez-Garcia, Maximilian Ilse, Daniel~C Castro, Benedikt Boecking, Harshita Sharma, Kenza Bouzid, Anja Thieme, et~al.
\newblock Learning to exploit temporal structure for biomedical vision-language processing.
\newblock In {\em Proceedings of the IEEE/CVF Conference on Computer Vision and Pattern Recognition}, pages 15016--15027, 2023.

\bibitem{beltagy_scibert_2019}
Iz Beltagy, Kyle Lo, and Arman Cohan.
\newblock {SciBERT}: {A} {Pretrained} {Language} {Model} for {Scientific} {Text}.
\newblock In {\em {EMNLP}}, 2019.

\bibitem{brown2020language}
Tom Brown, Benjamin Mann, Nick Ryder, Melanie Subbiah, Jared~D Kaplan, Prafulla Dhariwal, Arvind Neelakantan, Pranav Shyam, Girish Sastry, Amanda Askell, et~al.
\newblock Language models are few-shot learners.
\newblock {\em Advances in neural information processing systems}, 33:1877--1901, 2020.

\bibitem{chauhan_joint_2020}
Geeticka Chauhan, Ruizhi Liao, William Wells, Jacob Andreas, Xin Wang, Seth Berkowitz, Steven Horng, Peter Szolovits, and Polina Golland.
\newblock Joint {Modeling} of {Chest} {Radiographs} and {Radiology} {Reports} for {Pulmonary} {Edema} {Assessment}.
\newblock In {\em {MICCAI}}, 2020.

\bibitem{chen_simple_2020}
Ting Chen, Simon Kornblith, Mohammad Norouzi, and Geoffrey Hinton.
\newblock A {Simple} {Framework} for {Contrastive} {Learning} of {Visual} {Representations}.
\newblock In {\em Proceedings of the 37th {International} {Conference} on {Machine} {Learning}}, 2020.

\bibitem{chen_generating_2022}
Zhihong Chen, Yan Song, Tsung-Hui Chang, and Xiang Wan.
\newblock Generating radiology reports via memory-driven transformer, 2022.

\bibitem{cheng2023prior}
Pujin Cheng, Li Lin, Junyan Lyu, Yijin Huang, Wenhan Luo, and Xiaoying Tang.
\newblock Prior: Prototype representation joint learning from medical images and reports.
\newblock In {\em Proceedings of the IEEE/CVF International Conference on Computer Vision}, pages 21361--21371, 2023.

\bibitem{chowdhery2022palmscalinglanguagemodeling}
Aakanksha Chowdhery, Sharan Narang, Jacob Devlin, Maarten Bosma, Gaurav Mishra, Adam Roberts, Paul Barham, Hyung~Won Chung, Charles Sutton, Sebastian Gehrmann, Parker Schuh, Kensen Shi, Sasha Tsvyashchenko, Joshua Maynez, Abhishek Rao, Parker Barnes, Yi Tay, Noam Shazeer, Vinodkumar Prabhakaran, Emily Reif, Nan Du, Ben Hutchinson, Reiner Pope, James Bradbury, Jacob Austin, Michael Isard, Guy Gur-Ari, Pengcheng Yin, Toju Duke, Anselm Levskaya, Sanjay Ghemawat, Sunipa Dev, Henryk Michalewski, Xavier Garcia, Vedant Misra, Kevin Robinson, Liam Fedus, Denny Zhou, Daphne Ippolito, David Luan, Hyeontaek Lim, Barret Zoph, Alexander Spiridonov, Ryan Sepassi, David Dohan, Shivani Agrawal, Mark Omernick, Andrew~M. Dai, Thanumalayan~Sankaranarayana Pillai, Marie Pellat, Aitor Lewkowycz, Erica Moreira, Rewon Child, Oleksandr Polozov, Katherine Lee, Zongwei Zhou, Xuezhi Wang, Brennan Saeta, Mark Diaz, Orhan Firat, Michele Catasta, Jason Wei, Kathy Meier-Hellstern, Douglas Eck, Jeff Dean, Slav Petrov, and Noah Fiedel.
\newblock Palm: Scaling language modeling with pathways, 2022.

\bibitem{cornia_meshed-memory_2020}
Marcella Cornia, Matteo Stefanini, Lorenzo Baraldi, and Rita Cucchiara.
\newblock Meshed-memory transformer for image captioning, 2020.

\bibitem{desai_virtex_2021}
Karan Desai and Justin Johnson.
\newblock {VirTex}: {Learning} {Visual} {Representations} from {Textual} {Annotations}.
\newblock In {\em {CVPR}}, 2021.

\bibitem{dosovitskiy2021an}
Alexey Dosovitskiy, Lucas Beyer, Alexander Kolesnikov, Dirk Weissenborn, Xiaohua Zhai, Thomas Unterthiner, Mostafa Dehghani, Matthias Minderer, Georg Heigold, Sylvain Gelly, Jakob Uszkoreit, and Neil Houlsby.
\newblock An image is worth 16x16 words: Transformers for image recognition at scale.
\newblock In {\em {ICLR}}, 2021.

\bibitem{endo_retrieval-based_2021}
Mark Endo, Rayan Krishnan, Viswesh Krishna, Andrew~Y. Ng, and Pranav Rajpurkar.
\newblock Retrieval-{Based} {Chest} {X}-{Ray} {Report} {Generation} {Using} a {Pre}-trained {Contrastive} {Language}-{Image} {Model}.
\newblock In {\em Proceedings of {Machine} {Learning} for {Health}}, 2021.

\bibitem{he_deep_2015}
Kaiming He, Xiangyu Zhang, Shaoqing Ren, and Jian Sun.
\newblock Deep {Residual} {Learning} for {Image} {Recognition}, Dec. 2015.
\newblock arXiv:1512.03385 [cs].

\bibitem{holtzman_curious_2020}
Ari Holtzman, Jan Buys, Li Du, Maxwell Forbes, and Yejin Choi.
\newblock The curious case of neural text degeneration.
\newblock In {\em {ICLR}}, 2020.

\bibitem{horng_deep_2021}
Steven Horng, Ruizhi Liao, Xin Wang, Sandeep Dalal, Polina Golland, and Seth~J. Berkowitz.
\newblock Deep learning to quantify pulmonary edema in chest radiographs.
\newblock {\em Radiology: Artificial Intelligence}, 2021.

\bibitem{huang_gloria_2021}
Shih-Cheng Huang, Liyue Shen, Matthew~P. Lungren, and Serena Yeung.
\newblock {GLoRIA}: {A} {Multimodal} {Global}-{Local} {Representation} {Learning} {Framework} for {Label}-efficient {Medical} {Image} {Recognition}.
\newblock In {\em {ICCV}}, 2021.

\bibitem{huynh2022boosting}
Tri Huynh, Simon Kornblith, Matthew~R Walter, Michael Maire, and Maryam Khademi.
\newblock Boosting contrastive self-supervised learning with false negative cancellation.
\newblock In {\em Proceedings of the IEEE/CVF winter conference on applications of computer vision}, pages 2785--2795, 2022.

\bibitem{irvin_chexpert_2019}
Jeremy Irvin, Pranav Rajpurkar, Michael Ko, Yifan Yu, Silviana Ciurea-Ilcus, Chris Chute, Henrik Marklund, Behzad Haghgoo, Robyn Ball, Katie Shpanskaya, Jayne Seekins, David~A. Mong, Safwan~S. Halabi, Jesse~K. Sandberg, Ricky Jones, David~B. Larson, Curtis~P. Langlotz, Bhavik~N. Patel, Matthew~P. Lungren, and Andrew~Y. Ng.
\newblock {CheXpert}: {A} {Large} {Chest} {Radiograph} {Dataset} with {Uncertainty} {Labels} and {Expert} {Comparison}.
\newblock {\em Proceedings of the AAAI Conference on Artificial Intelligence}, 2019.

\bibitem{jing_automatic_2018}
Baoyu Jing, Pengtao Xie, and Eric Xing.
\newblock On the {Automatic} {Generation} of {Medical} {Imaging} {Reports}.
\newblock In {\em Proceedings of the 56th {Annual} {Meeting} of the {Association} for {Computational} {Linguistics}}, 2018.

\bibitem{johnson_mimic-cxr_2019}
Alistair E.~W. Johnson, Tom~J. Pollard, Seth~J. Berkowitz, Nathaniel~R. Greenbaum, Matthew~P. Lungren, Chih-ying Deng, Roger~G. Mark, and Steven Horng.
\newblock {MIMIC}-{CXR}, a de-identified publicly available database of chest radiographs with free-text reports.
\newblock {\em Scientific Data}, 2019.

\bibitem{johnson_mimic-cxr-jpg_2019}
Alistair E.~W. Johnson, Tom~J. Pollard, Nathaniel~R. Greenbaum, Matthew~P. Lungren, Chih ying Deng, Yifan Peng, Zhiyong Lu, Roger~G. Mark, Seth~J. Berkowitz, and Steven Horng.
\newblock Mimic-cxr-jpg, a large publicly available database of labeled chest radiographs, 2019.

\bibitem{kojima2022large}
Takeshi Kojima, Shixiang~Shane Gu, Machel Reid, Yutaka Matsuo, and Yusuke Iwasawa.
\newblock Large language models are zero-shot reasoners.
\newblock {\em Advances in neural information processing systems}, 35:22199--22213, 2022.

\bibitem{kudo_sentencepiece_2018}
Taku Kudo and John Richardson.
\newblock {SentencePiece}: {A} simple and language independent subword tokenizer and detokenizer for {Neural} {Text} {Processing}.
\newblock In {\em {EMNLP}: {System} {Demonstrations}}, 2018.

\bibitem{li_unicoder-vl_2020}
Gen Li, Nan Duan, Yuejian Fang, Ming Gong, and Daxin Jiang.
\newblock Unicoder-{VL}: {A} {Universal} {Encoder} for {Vision} and {Language} by {Cross}-{Modal} {Pre}-{Training}.
\newblock {\em Proceedings of the AAAI Conference on Artificial Intelligence}, 2020.

\bibitem{li_hybrid_2018}
Yuan Li, Xiaodan Liang, Zhiting Hu, and Eric~P Xing.
\newblock Hybrid {Retrieval}-{Generation} {Reinforced} {Agent} for {Medical} {Image} {Report} {Generation}.
\newblock In {\em {NeurIPS}}, 2018.

\bibitem{liao_multimodal_2021}
Ruizhi Liao, Daniel Moyer, Miriam Cha, Keegan Quigley, Seth Berkowitz, Steven Horng, Polina Golland, and William~M. Wells.
\newblock Multimodal {Representation} {Learning} via {Maximization} of {Local} {Mutual} {Information}.
\newblock In {\em {MICCAI}}, 2021.

\bibitem{lin_microsoft_2014}
Tsung-Yi Lin, Michael Maire, Serge Belongie, James Hays, Pietro Perona, Deva Ramanan, Piotr Dollár, and C.~Lawrence Zitnick.
\newblock Microsoft {COCO}: {Common} {Objects} in {Context}.
\newblock In {\em {ECCV}}, 2014.

\bibitem{lu_vilbert_2019}
Jiasen Lu, Dhruv Batra, Devi Parikh, and Stefan Lee.
\newblock {ViLBERT}: {Pretraining} {Task}-{Agnostic} {Visiolinguistic} {Representations} for {Vision}-and-{Language} {Tasks}.
\newblock In {\em {NeurIPS}}, 2019.

\bibitem{mishra_improving_2023}
Aakash Mishra, Rajat Mittal, Christy Jestin, Kostas Tingos, and Pranav Rajpurkar.
\newblock Improving zero-shot detection of low prevalence chest pathologies using domain pre-trained language models, 2023.

\bibitem{miura-etal-2021-improving}
Yasuhide Miura, Yuhao Zhang, Emily Tsai, Curtis Langlotz, and Dan Jurafsky.
\newblock Improving factual completeness and consistency of image-to-text radiology report generation.
\newblock In {\em Proceedings of the 2021 Conference of the North American Chapter of the Association for Computational Linguistics: Human Language Technologies}, 2021.

\bibitem{nicolson2023improving}
Aaron Nicolson, Jason Dowling, and Bevan Koopman.
\newblock Improving chest x-ray report generation by leveraging warm starting.
\newblock {\em Artificial intelligence in medicine}, 144:102633, 2023.

\bibitem{pavlova_covidx_2022}
Maya Pavlova, Tia Tuinstra, Hossein Aboutalebi, Andy Zhao, Hayden Gunraj, and Alexander Wong.
\newblock {COVIDx} {CXR}-3: {A} {Large}-{Scale}, {Open}-{Source} {Benchmark} {Dataset} of {Chest} {X}-ray {Images} for {Computer}-{Aided} {COVID}-19 {Diagnostics}, Nov. 2022.
\newblock arXiv:2206.03671 [cs, eess].

\bibitem{pino_clinically_2021}
Pablo Pino, Denis Parra, Cecilia Besa, and Claudio Lagos.
\newblock Clinically {Correct} {Report} {Generation} from {Chest} {X}-{Rays} {Using} {Templates}.
\newblock In {\em Machine {Learning} in {Medical} {Imaging}}. Springer International Publishing, 2021.

\bibitem{press2016using}
Ofir Press and Lior Wolf.
\newblock Using the output embedding to improve language models.
\newblock {\em arXiv preprint arXiv:1608.05859}, 2016.

\bibitem{radford_learning_2021}
Alec Radford, Jong~Wook Kim, Chris Hallacy, Aditya Ramesh, Gabriel Goh, Sandhini Agarwal, Girish Sastry, Amanda Askell, Pamela Mishkin, Jack Clark, Gretchen Krueger, and Ilya Sutskever.
\newblock Learning transferable visual models from natural language supervision, 2021.

\bibitem{radford2018improving}
Alec Radford, Karthik Narasimhan, Tim Salimans, Ilya Sutskever, et~al.
\newblock Improving language understanding by generative pre-training.
\newblock 2018.

\bibitem{raffel2020exploring}
Colin Raffel, Noam Shazeer, Adam Roberts, Katherine Lee, Sharan Narang, Michael Matena, Yanqi Zhou, Wei Li, and Peter~J Liu.
\newblock Exploring the limits of transfer learning with a unified text-to-text transformer.
\newblock {\em Journal of machine learning research}, 21(140):1--67, 2020.

\bibitem{ramesh_improving_2022}
Vignav Ramesh, Nathan~A. Chi, and Pranav Rajpurkar.
\newblock Improving {Radiology} {Report} {Generation} {Systems} by {Removing} {Hallucinated} {References} to {Non}-existent {Priors}.
\newblock In {\em Proceedings of the 2nd {Machine} {Learning} for {Health} symposium}, 2022.

\bibitem{rohrbach2018object}
Anna Rohrbach, Lisa~Anne Hendricks, Kaylee Burns, Trevor Darrell, and Kate Saenko.
\newblock Object hallucination in image captioning.
\newblock {\em arXiv preprint arXiv:1809.02156}, 2018.

\bibitem{seyyed2021underdiagnosis}
Laleh Seyyed-Kalantari, Haoran Zhang, Matthew~BA McDermott, Irene~Y Chen, and Marzyeh Ghassemi.
\newblock Underdiagnosis bias of artificial intelligence algorithms applied to chest radiographs in under-served patient populations.
\newblock {\em Nature medicine}, 27(12):2176--2182, 2021.

\bibitem{shih_augmenting_2019}
George Shih, Carol~C. Wu, Safwan~S. Halabi, Marc~D. Kohli, Luciano~M. Prevedello, Tessa~S. Cook, Arjun Sharma, Judith~K. Amorosa, Veronica Arteaga, Maya Galperin-Aizenberg, Ritu~R. Gill, Myrna~C.B. Godoy, Stephen Hobbs, Jean Jeudy, Archana Laroia, Palmi~N. Shah, Dharshan Vummidi, Kavitha Yaddanapudi, and Anouk Stein.
\newblock Augmenting the national institutes of health chest radiograph dataset with expert annotations of possible pneumonia.
\newblock {\em Radiology: Artificial Intelligence}, 2019.

\bibitem{smit2020chexbert}
Akshay Smit, Saahil Jain, Pranav Rajpurkar, Anuj Pareek, Andrew~Y Ng, and Matthew~P Lungren.
\newblock Chexbert: combining automatic labelers and expert annotations for accurate radiology report labeling using bert.
\newblock {\em arXiv preprint arXiv:2004.09167}, 2020.

\bibitem{stefanini_show_2023}
M. Stefanini, M. Cornia, L. Baraldi, S. Cascianelli, G. Fiameni, and R. Cucchiara.
\newblock From show to tell: A survey on deep learning-based image captioning.
\newblock {\em IEEE Transactions on Pattern Analysis \& Machine Intelligence}, 2023.

\bibitem{tiu_expert-level_2022}
Ekin Tiu, Ellie Talius, Pujan Patel, Curtis~P. Langlotz, Andrew~Y. Ng, and Pranav Rajpurkar.
\newblock Expert-level detection of pathologies from unannotated chest {X}-ray images via self-supervised learning.
\newblock {\em Nature Biomedical Engineering}, 2022.

\bibitem{tu2024towards}
Tao Tu, Shekoofeh Azizi, Danny Driess, Mike Schaekermann, Mohamed Amin, Pi-Chuan Chang, Andrew Carroll, Charles Lau, Ryutaro Tanno, Ira Ktena, et~al.
\newblock Towards generalist biomedical ai.
\newblock {\em NEJM AI}, 1(3):AIoa2300138, 2024.

\bibitem{vaswani_attention_2017}
Ashish Vaswani, Noam Shazeer, Niki Parmar, Jakob Uszkoreit, Llion Jones, Aidan~N Gomez, Lukasz Kaiser, and Illia Polosukhin.
\newblock Attention is {All} you {Need}.
\newblock In {\em {NeurIPS}}, 2017.

\bibitem{vinyals_show_2015}
Oriol Vinyals, Alexander Toshev, Samy Bengio, and Dumitru Erhan.
\newblock Show and tell: A neural image caption generator, 2015.

\bibitem{wang2022multi}
Fuying Wang, Yuyin Zhou, Shujun Wang, Varut Vardhanabhuti, and Lequan Yu.
\newblock Multi-granularity cross-modal alignment for generalized medical visual representation learning.
\newblock {\em Advances in Neural Information Processing Systems}, 35:33536--33549, 2022.

\bibitem{wang_chestx-ray8_2017}
Xiaosong Wang, Yifan Peng, Le Lu, Zhiyong Lu, Mohammadhadi Bagheri, and Ronald~M Summers.
\newblock {ChestX}-ray8: {Hospital}-{Scale} {Chest} {X}-{Ray} {Database} and {Benchmarks} on {Weakly}-{Supervised} {Classification} and {Localization} of {Common} {Thorax} {Diseases}.
\newblock In {\em {CVPR}}, 2017.

\bibitem{wang_tienet_2018}
Xiaosong Wang, Yifan Peng, Le Lu, Zhiyong Lu, and Ronald~M. Summers.
\newblock {TieNet}: {Text}-{Image} {Embedding} {Network} for {Common} {Thorax} {Disease} {Classification} and {Reporting} in {Chest} {X}-rays.
\newblock In {\em {CVPR}}, 2018.

\bibitem{yan2022clinical}
Bin Yan and Mingtao Pei.
\newblock Clinical-bert: Vision-language pre-training for radiograph diagnosis and reports generation.
\newblock In {\em Proceedings of the AAAI Conference on Artificial Intelligence}, volume~36, pages 2982--2990, 2022.

\bibitem{yang2024limits}
Yuzhe Yang, Haoran Zhang, Judy~W Gichoya, Dina Katabi, and Marzyeh Ghassemi.
\newblock The limits of fair medical imaging ai in real-world generalization.
\newblock {\em Nature Medicine}, pages 1--11, 2024.

\bibitem{youden1950index}
William~J Youden.
\newblock Index for rating diagnostic tests.
\newblock {\em Cancer}, 3(1):32--35, 1950.

\bibitem{yu_evaluating_2022}
Feiyang Yu, Mark Endo, Rayan Krishnan, Ian Pan, Andy Tsai, Eduardo~Pontes Reis, Eduardo Kaiser Ururahy~Nunes Fonseca, Henrique Min~Ho Lee, Zahra Shakeri~Hossein Abad, Andrew~Y. Ng, Curtis~P. Langlotz, Vasantha~Kumar Venugopal, and Pranav Rajpurkar.
\newblock Evaluating progress in automatic chest x-ray radiology report generation.
\newblock {\em medRxiv}, 2022.

\bibitem{yu_coca_2022}
Jiahui Yu, Zirui Wang, Vijay Vasudevan, Legg Yeung, Mojtaba Seyedhosseini, and Yonghui Wu.
\newblock Coca: Contrastive captioners are image-text foundation models, 2022.

\bibitem{zhang_lookahead_2019}
Michael Zhang, James Lucas, Jimmy Ba, and Geoffrey~E Hinton.
\newblock Lookahead {Optimizer}: k steps forward, 1 step back.
\newblock In {\em {NeurIPS}}, 2019.

\bibitem{zhang_bertscore_2019}
Tianyi Zhang, Varsha Kishore, Felix Wu, Kilian~Q. Weinberger, and Yoav Artzi.
\newblock Bertscore: Evaluating text generation with bert.
\newblock In {\em {ICLR}}, 2020.

\bibitem{zhang_knowledge-enhanced_2023}
Xiaoman Zhang, Chaoyi Wu, Ya Zhang, Weidi Xie, and Yanfeng Wang.
\newblock Knowledge-enhanced visual-language pre-training on chest radiology images.
\newblock {\em Nat Commun}, 14(1), July 2023.

\bibitem{zhang_contrastive_2022}
Yuhao Zhang, Hang Jiang, Yasuhide Miura, Christopher~D. Manning, and Curtis~P. Langlotz.
\newblock Contrastive learning of medical visual representations from paired images and text, 2022.

\bibitem{zhou_unified_2020}
Luowei Zhou, Hamid Palangi, Lei Zhang, Houdong Hu, Jason Corso, and Jianfeng Gao.
\newblock Unified {Vision}-{Language} {Pre}-{Training} for {Image} {Captioning} and {VQA}.
\newblock {\em Proceedings of the AAAI Conference on Artificial Intelligence}, 2020.

\end{thebibliography}
}

\clearpage
\appendix
\label{sec:app}
\section{Additional Implementation Details}
\label{sec:extra_implementation_details}
The transformer decoder in RadTex is implemented following VirTex \cite{desai_virtex_2021}, with a hidden dimension size of 2048, 32 attention heads, and a feedforward size of 8192. Each layer of the transformer decoder includes dropout ($p=0.1$) subsequent to layer normalization. 

During training, images are transformed with augmentations before passed to the model. All images are normalized by the MIMIC CXR mean and standard deviation values. Images are then randomly rotated on the range $[-20^\circ, +20^\circ]$, randomly cropped to a square, resized to $224\times224$, and then randomly rotated to $\{0^\circ, 90^\circ, 180^\circ, 270^\circ\}$. This sequence of random rotations and cropping preserves minimal image padding, while adding small rotations to adjust for imaging differences, and $90^\circ$ rotations to adjust for recording errors. These augmentations are specifically designed for the radiology domain.

All models are pretrained with an SGD optimizer with Lookahead ($k=5$, $\alpha=0.5$, as in \cite{zhang_lookahead_2019}). During pretraining, a linear warmup of the learning rate (5\% of total steps) is followed by cosine decay, with maximum LR of 0.4 and 0.002 for visual encoder and transformer decoder, respectively. Models are pretrained on 8 Nvidia V100 GPUs with an effective batch size of 512 samples. For visual downstream tasks, we train a classification head for the visual encoder on a single V100 GPU with a learning rate of 0.02 and batch size of 16, reducing LR by half if validation loss has not decreased after 5 epochs.

\subsection{Autoregressive Prompted Captioning Details}
\label{app:prompted-captioning}
Early experiments found nucleus sampling produced more specific and diverse findings than beam search, so it is used as our default sampling strategy. 
We evaluate RadTex radiology report generation (RRG) under three configurations: \textit{Unprompted}, \textit{Prompted}, and \textit{Iterative Prompted}, as defined in \cref{sec:methods_captioning}. Here, we provide additional details on our approach to prompted captioning.

Recall that the initial token sequence $P$ can optionally be non-empty during autoregressive token prediction and nucleus sampling. Non-empty $P$ conditions generation on prior tokens, encouraging the model to align its token prediction with this text during generation. We denote the tokenization of input text with tokenizer $\phi_{TOK}$. Special token embeddings $w^*$ stop generation when reached, so adding $\phi_{TOK}(\texttt{"."})$ as a special token can effectively create a sentence-length output for a single prompt and allow for structured report generation from multiple prompts. \cref{alg:captioning} describes forward (left-to-right) captioning, but backward captioning is also possible by modifying the $w^*$ special token embeddings, initializing $w$ with \texttt{"[SEP]"}, and changing $\phi_f \rightarrow \phi_b$.
\begin{algorithm}
\small
\caption{Autoregressive Prompted Captioning}\label{alg:captioning}
\hspace*{\algorithmicindent} \textbf{Input:} Prompt $P$; Sampling Strategy $\eta$; Image $I$ \\
\hspace*{\algorithmicindent} \textbf{Output:} Caption tokens $w$ 
\begin{algorithmic}[1]
\State $x_{vis} \gets \theta(I)$ \Comment{Visual Embedding}

\State $w \gets [\phi_{TOK}(\texttt{[SOS]}), \phi_{TOK}(P)]$ \Comment{Tokenize}
\State $w^* \gets [\phi_{TOK}(\texttt{[SEP]})]$ \Comment{Stop Tokens}
\State $t \gets $ length($w$) + 1
\While {$t < \text{max caption length}$}
    \State $x_{text} \gets \phi_{EMB}(w)$ \Comment{Textual Embedding}
    \State $l \gets \phi_f(x_{vis}, x_{text})$ \Comment{Token Logits}
    \State $w_t \gets \eta(l) \in D$ \Comment{New token}
    \State $w[t] \gets w_t$
    \State $t \gets t+1$
    \If {$w_t \in w^*$}{
        break}
    \EndIf
\EndWhile
\end{algorithmic}
\end{algorithm}

For prompted captioning, we use the stop tokens to limit generation to a single sentence, and then concatenate sentences from multiple prompts to form the \textit{Prompted} report. We use this approach for our \textit{Iterative Prompted} reports as well, but allow each sentence to be modified via backward captioning. 

\subsection{VirTex Comparison}
When comparing RadTex results to other models on the CheXpert competition classification task in \cref{tab:chexpert_partial}, we report performance for VirTex/C+M. This VirTex model is identical to the architecture presented in Desai and Johnson \cite{desai_virtex_2021}, but we re-pretrain on MS-COCO, followed by MIMIC-CXR, rather than using VirTex weights. Crucially, we keep the context length at 30 tokens for each of these stages of pretraining. However, because the MIMIC-CXR reports typically span more than 30 tokens, we configure the dataloader to randomly select a sentence from the report when loading a batch. This randomization allows the entire report to be seen during pretraining with many epochs, giving models with shorter context-length a better chance to perform competitively with longer context-length models.

\section{Visual Encoder Results}
\label{app:vision_encoder_results}

\cref{tab:visual_experiments} shows the full performance of RadTex compared to baselines on our four main downstream tasks, including standard error. AUCPR is reported alongside AUC, to better study performance on negative examples. We found that models with the highest AUC also had the highest AUCPR.

\begin{table*}[ht]
\centering
    \begin{tabular}{l|rr|rr|rr|r}
    \toprule
    {} &           \multicolumn{2}{c}{CheXpert} &        \multicolumn{2}{c}{RSNA Pneumonia} &             \multicolumn{2}{c}{COVIDx} &   Edema Sev. \\
    {} &            \multicolumn{1}{c}{\tiny{AUC (\%)}} & \multicolumn{1}{c}{\tiny{AUCPR (\%)}}    &    \multicolumn{1}{c}{\tiny{AUC (\%)}} & \multicolumn{1}{c}{\tiny{AUCPR (\%)}}  &            \multicolumn{1}{c}{\tiny{AUC (\%)}} & \multicolumn{1}{c}{\tiny{AUCPR (\%)}}    & \tiny{macro-F1} \\
\midrule
RadTex      &  $\mathbf{89.44}_{\pm0.06}$ &  $\underline{\mathbf{61.42}}_{\pm0.04}$ &  $88.56_{\pm0.02}$ &  $70.52_{\pm0.05}$ &  $\underline{\mathbf{98.14}}_{\pm0.18}$ &  $\underline{\mathbf{98.06}}_{\pm0.19}$ &  $\underline{\mathbf{0.59}}_{\pm0.00}$ \\
CheXzero    &  $89.04_{\pm0.21}$ &  $59.59_{\pm0.55}$ &  $\mathbf{88.70}_{\pm0.06}$ &  $\underline{\mathbf{71.58}}_{\pm0.21}$ &  $96.00_{\pm0.05}$ &  $96.48_{\pm0.04}$ &  $0.51_{\pm0.03}$ \\
BiomedCLIP &  $83.13_{\pm0.28}$ &  $52.15_{\pm0.57}$ &  $88.12_{\pm0.01}$ &  $70.38_{\pm0.05}$ &  $97.04_{\pm0.02}$ &  $97.20_{\pm0.02}$ &  $0.56_{\pm0.01}$ \\
CLIP        &  $73.34_{\pm0.13}$ &  $35.74_{\pm0.26}$ &  $81.57_{\pm0.02}$ &  $55.79_{\pm0.08}$ &  $97.50_{\pm0.08}$ &  $97.73_{\pm0.07}$ &  $0.31_{\pm0.02}$ \\
\midrule
Supervised  &  $85.01_{\pm2.73}$ &  $54.62_{\pm5.47}$ &  $86.76_{\pm0.20}$ &  $67.60_{\pm0.59}$ &  $\mathbf{99.37}_{\pm0.27}$ &  $\mathbf{99.57}_{\pm0.14}$ &  $0.45_{\pm0.01}$ \\
\bottomrule
    \end{tabular}
    \caption{Results of RadTex and compared contrastive pretraining methods on four downstream tasks. AUC and AUCPR are reported for binary classification datasets and also denote macro-AUC and macro-AUCPR for five-label CheXpert competition dataset. Edema Sev. displays the macro-F1 score averaged across the four ordinal severity grades. Supervised method is fine-tuned end-to-end, while other methods have a frozen pretrained visual backbone and a trainable linear layer. Best pretrained result for each metric is \textbf{bolded}, as well as the Supervised result if it exceeds pretraining methods. Results that pass a one-sided t-test for significance at $p=0.05$ are \underline{\textbf{underlined}}. $\text{Mean}_{\pm \text{SE}}$ over three random trials reported.}
    \label{tab:visual_experiments}

\end{table*}



\section{Radiology Report Generation}
In this section, we provide additional information on the performance values drawn from the literature and the experiments used to test generated report quality. 

\subsection{Literature Comparisons}

In Table~\ref{tab:captioning}, we re-print results reported in existing literature. Here, we detail the listed values for R2Gen, $\mathcal{M}^2$ Trans, CXR-RePaiR, and MedPaLM M. 

For metrics BERTscore, CheXbert Sim., and RadGraph F1, we take numbers directly from \cite{yu_evaluating_2022} (Supplementary Table 4). These listed scores were based on comparisons to \textit{Impression} section as ground truth, so we only compare to \textit{Impression} when computing our scores using their code. Their scores also represent the average of test set \textit{studies}, rather than images (multiple radiographs can be associated with a single study). To select the X-ray used to generate a report for a given study, a PA or AP view image is randomly selected from associated examples. If no PA or AP view is available, a random example from the study is selected. We follow the same selection process when computing these three metrics and comparing to their results.

For metrics BLEU-2 and CheXpert macro-F1, we compute performance on the full number of test set image-report pairs (5,159), and compute BLEU-2 by comparing to the \textit{Findings} and \textit{Impression} sections of the report. We draw numbers for R2Gen from Chen \etal \cite{chen_generating_2022} (Table 2, Base model + relational memory + memory-driven conditional layer normalization), and numbers for CXR-RePaiR from Endo \etal \cite{endo_retrieval-based_2021} (Table 1, CXR-RePaiR-2). For M2 Trans, we compute the macro-F1 score based on the per-pathology F1 scores in Miura \etal \cite{miura-etal-2021-improving}, Tables 6 and 7 (NLL loss with BERTscore and Entailing Entity Match Reward). BLEU-2 score was not reported  for M2 Trans.

MedPaLM M reports their clinical efficacy scores in Tu \etal \cite{tu2024towards}, Table 2. They do not report BLEU-2 or BERTScore, and do not provide code or results for further analysis. MedPaLM M predicts the \textit{Findings} section of the target report, but also includes the \textit{Indication} section as context during generation. This additional context may give the MedPaLM M indications about the image contents that boost it's clinical efficacy and reduce reliance on the visual encoder for extracting information from the images.

\subsection{Per-Pathology Report Quality}
\label{app:per-pathology_report_quality}

\begin{table*}[h]
\centering
\begin{tabular}{lp{0.60cm}p{0.60cm}p{0.75cm}|p{0.60cm}p{0.60cm}p{0.75cm}|p{0.60cm}p{0.60cm}p{0.75cm}|p{0.60cm}p{0.60cm}p{0.75cm}|p{0.6cm}}
\toprule
{} & \multicolumn{3}{c}{Random Retrieval} & \multicolumn{3}{c}{Unprompted} & \multicolumn{3}{c}{Prompted} & \multicolumn{3}{c}{Iterative Prompted} & Train Freq. \\
 &                P &     R &    F1 &          P &     R &    F1 &        P &     R &    F1 &                  P &     R & \multicolumn{2}{l}{F1} \\
\midrule
Atelectasis                &            0.234 & 0.271 & 0.251 &      0.347 & 0.423 & 0.381 &    0.394 & 0.729 & 0.512 &              0.390 & 0.749 & \textbf{0.513} &              0.253 \\
Cardiomegaly               &            0.276 & 0.334 & 0.302 &      0.369 & 0.501 & 0.425 &    0.440 & 0.788 & \textbf{0.565} &              0.430 & 0.786 & 0.556 &              0.292 \\
Consolidation              &            0.089 & 0.082 & 0.086 &      0.151 & 0.152 & 0.152 &    0.188 & 0.514 & \textbf{0.275} &              0.185 & 0.524 & 0.274 &              0.101 \\
Edema                      &            0.265 & 0.192 & 0.223 &      0.468 & 0.422 & 0.444 &    0.475 & 0.789 & \textbf{0.593} &              0.472 & 0.774 & 0.586 &              0.287 \\
Enlarged C. &            0.088 & 0.179 & 0.118 &      0.086 & 0.220 & \textbf{0.124} &    0.099 & 0.070 & 0.082 &              0.087 & 0.070 & 0.078 &              0.095 \\
Fracture                   &            0.019 & 0.023 & 0.021 &      0.085 & 0.069 & \textbf{0.076} &    0.031 & 0.006 & 0.010 &              0.023 & 0.006 & 0.009 &              0.026 \\
Lung Lesion                &            0.042 & 0.038 & 0.040 &      0.086 & 0.051 & \textbf{0.064} &    0.057 & 0.025 & 0.035 &              0.074 & 0.046 & 0.057 &              0.036 \\
Lung Opacity               &            0.324 & 0.319 & 0.322 &      0.398 & 0.391 & 0.394 &    0.450 & 0.638 & 0.528 &              0.443 & 0.669 & \textbf{0.533} &              0.255 \\
Pleural Effusion           &            0.346 & 0.314 & 0.329 &      0.543 & 0.579 & 0.560 &    0.636 & 0.761 & \textbf{0.693} &              0.613 & 0.770 & 0.683 &              0.382 \\
Pleural Other              &            0.057 & 0.034 & 0.043 &      0.049 & 0.028 & 0.035 &    0.079 & 0.034 & \textbf{0.048} &              0.055 & 0.021 & 0.030 &              0.013 \\
Pneumonia                  &            0.212 & 0.175 & 0.191 &      0.299 & 0.236 & 0.264 &    0.339 & 0.633 & \textbf{0.442} &              0.341 & 0.583 & 0.430 &              0.259 \\
Pneumothorax               &            0.022 & 0.041 & 0.028 &      0.154 & 0.275 & \textbf{0.197} &    0.065 & 0.901 & 0.122 &              0.085 & 0.632 & 0.149 &              0.236 \\
Support Devices            &            0.284 & 0.392 & 0.329 &      0.523 & 0.618 & 0.566 &    0.675 & 0.636 & \textbf{0.655} &              0.663 & 0.636 & 0.649 &              0.308 \\
No Finding                 &            0.186 & 0.197 & 0.192 &      0.396 & 0.338 & \textbf{0.365} &    0.562 & 0.227 & 0.323 &              0.535 & 0.247 & 0.338 &              0.334 \\
\midrule
Macro-Average & 0.174 &0.185 & 0.177& 0.282& 0.307& 0.289& 0.321& 0.482& \textbf{0.349}& 0.314& 0.465& \textbf{0.349}&--- \\
\bottomrule
\end{tabular}
\caption{CheXpert label scores for four different methods of captioning on the MIMIC-CXR test set, divided by pathology and averaged at the image-level. Training set frequency, as fraction of reports containing Chexpert-labeled pathology, is shown at the right. Enlarged C. stands for Enlarged Cardiomediastinum. Best F1 Scores for each pathology are \textbf{bolded.}}
\label{tab:per_pathology}

\end{table*}

In Table~\ref{tab:per_pathology}, we report the per-pathology breakdown of CheXpert labeling for Unprompted, Prompted, Iterative Prompted, and Random Retrieval captioning methods. For Random Retrieval, we randomly select a report from the training set to use as a baseline for report generation. Beyond CheXpert F1, Precision, and Recall scores, which represent measures of clinical efficacy, we also report the frequency of each pathology in the training dataset. This is computed as the fraction of total 222,750 non-empty ground truth reports that are labeled for a pathology by the Chexpert labeler.

\begin{figure*}[ht]
    \centering
    \includegraphics[width=0.8\textwidth]{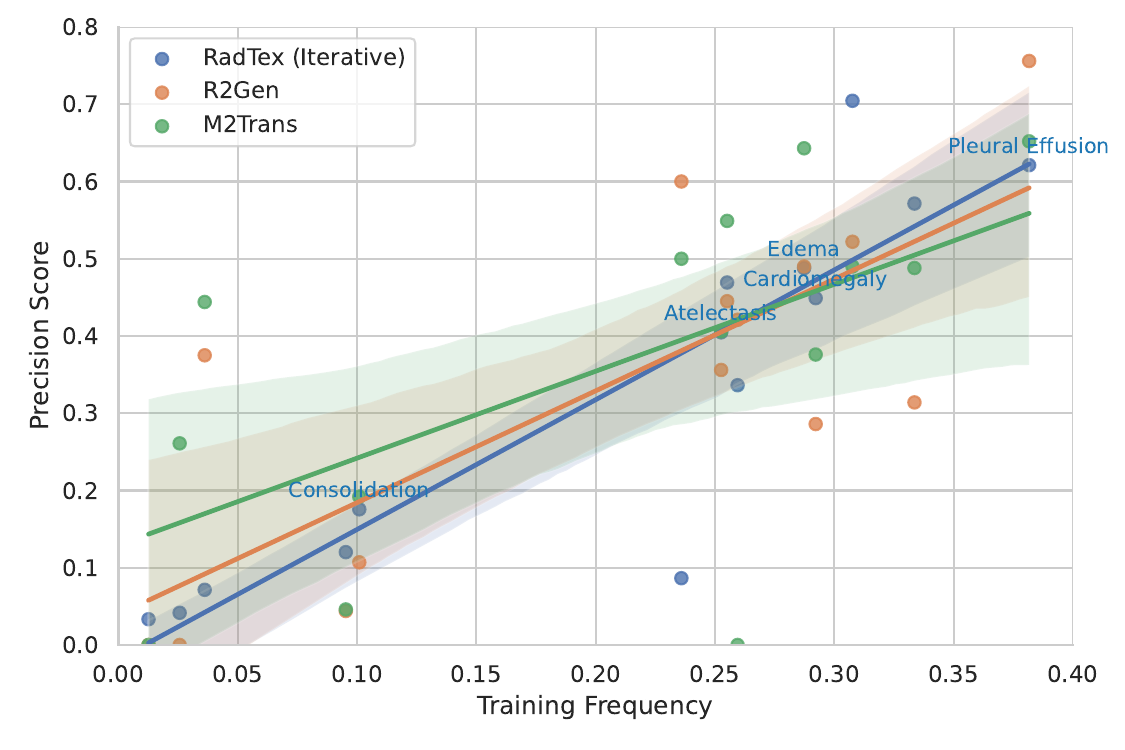}
    \caption{Report-level pathology precision on MIMIC-CXR test set vs frequency in training data, as a measure of pathology hallucination. $\mathcal{M}^2$ Trans and R2 Gen numbers from \cite{miura-etal-2021-improving}. Linear regressions and 95\% confidence intervals for each RRG method are shown. Pearson correlation coefficients of $r=0.89, 0.78,0.62$ for RadTex, R2Gen, and  $\mathcal{M}^2$ Trans, respectively. The CheXpert competition pathologies are labeled for RadTex.}
    \label{fig:per_pathology_captioning}
\end{figure*}

In Figure~\ref{fig:per_pathology_captioning}, we plot these results, showing a clear positive trend between the pathology precision score and training frequency. This trend holds for all four captioning methods, including Random Retrieval.

\subsection{Text Generation Examples}

Additional examples of radiology report captioning are shown in \cref{tab:extra_cap_examples}. When comparing \textit{Iterative Prompted} and \textit{Prompted} captioning results, we observe that iterative prompting does effectively allow for more detailed descriptions of pathologies. By backward captioning with a forward-prompted output, Iterative Prompting yields reports that use modifiers to better describe findings. For example, in Table~\ref{tab:extra_cap_examples}, Iterative Prompting allows for ``\textit{interstitial} edema..." (row 1), and ``\textit{minimal} atelectasis..." (row 3). We hypothesize that the ability to add modifiers in front of pathology descriptions may contribute to the higher RadGraph F1 score with Iterative Prompting (Table~\ref{tab:captioning}).

\begin{landscape}
\begin{table}
    \centering

    \small
         \begin{supertabular}{ c p{0.23\textwidth} p{0.27\linewidth} p{0.13\linewidth} p{0.16\linewidth}}
         \toprule
         Image & Ground Truth & Unprompted Report & Prompted Report & Iterative Prompted \\
         \midrule
              \raisebox{-0.9\totalheight}{\includegraphics[width=0.17\textwidth]{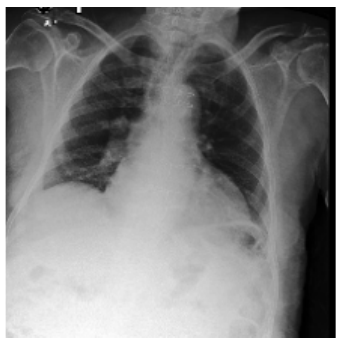}}
            &  No acute cardiopulmonary process.
    
            & There is mild cardiomegaly. The aorta is mildly tortuous. No focal consolidation, pleural effusion, or pneumothorax is seen. The cardiac and mediastinal silhouettes are stable. Old healed right rib fractures are present. Degenerative changes are seen along the thoracic spine. No acute cardiopulmonary process.
    
            & Edema is now moderate.
    
            & Interstitial edema is again seen in the right lung.
          \\ 
    
          \midrule
           \raisebox{-1.1\totalheight}{\includegraphics[width=0.17\textwidth]{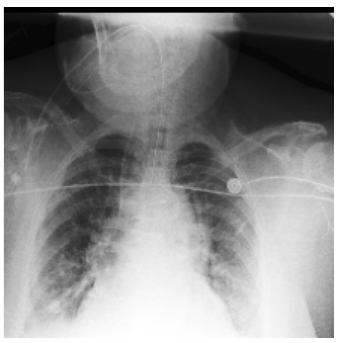}}
            &  As compared to the previous radiograph, the endotracheal tube, nasogastric tube and right internal jugular vein catheter are unchanged. The pre-existing pulmonary edema might have mildly improved, there is increased retrocardiac and right basal atelectasis. No pleural effusions. No major atelectasis.
          & 
          Mild pulmonary edema has worsened and has not decreased. Right lower lobe opacity has probably a combination of pleural effusion or alternatively lymphadenopathy. The catheter has been removed. The heart size is mildly enlarged but is normal. The pulmonary vasculature is not engorged. There is no pneumothorax or pleural effusion. No pneumothorax. Sternotomy wires are noted. Interval placement of a nasogastric tube whose tip number - up for its tip over number cm. number. Et tube ends within number cm above the carina. New right atrium. Number.
          &
          Edema has changed in the patient has worsened since. Support devices are unchanged in position.
          &  Enteric tube suggests mild pulmonary edema. Monitoring and support devices remain in unchanged position, right central venous catheter in place.
          \\ 
          
          \midrule
           \raisebox{-1.1\totalheight}{\includegraphics[width=0.17\textwidth]{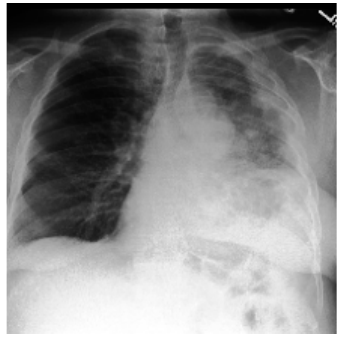}}
            & The patient has had a prior left lower lobectomy. Since the prior exam, nodular pleural thickening encasing the left lung has increased at the expense of aeration of the left lung with stable elevation of the left hemidiaphragm. Central adenopathy in the left hilus and adjacent mediastinum has also progressed. The right lung is clear. Cardiomediastinal silhouette is unchanged. Progression of left pleural and nodal metastases.
    
            & Pa and lateral chest views were provided. There is extensive areas of consolidation in the right lung with a left lung likely representing metastatic lesion. There is a moderate right pleural effusion. There is a large left pleural effusion. There is persistent left effusion. Degenerative changes noted in the thoracic spine, better visualized on the right. Cardiomediastinal silhouette appears similar. Bony structures are intact. The possibility of pneumonia.
    
            & Atelectasis at the left lung base. Consolidation in the left upper lobe is stable. Lung opacity is prominent. Pleural effusion is volume loss in the right lung. Pneumonia or is known between the left upper and has changed in diameter. Pneumothorax.
            & Minimal atelectasis at the right lung base. Airspace consolidation at the right upper lung is unchanged. Left lower lung opacity is concerning for pneumonia, but stable. Minimal right basilar loculated pleural effusion and possibly moderate on the left. Unchanged right basilar opacity concerning for pneumonia is noted within the left upper lobe. There is no pneumothorax have been stable.
          \\ 
          \bottomrule
          \end{supertabular}
          \caption{RadTex/C+M captioning examples in \textit{Unprompted}, \textit{Prompted}, and \textit{Iterative Prompted} captioning setups with corresponding CXR images}
          \label{tab:extra_cap_examples}
      \end{table}
\end{landscape}




\end{document}